\documentclass{article}

\usepackage{arxiv}

\usepackage[utf8]{inputenc} 
\usepackage[T1]{fontenc}    
\usepackage{hyperref}       
\usepackage{url}            
\usepackage{booktabs}       
\usepackage{amsfonts}       
\usepackage{nicefrac}       
\usepackage{microtype}      
\usepackage{lipsum}		
\usepackage{graphicx}
\usepackage{natbib}
\usepackage{doi}

\title{4Weed Dataset: Annotated Imagery Weeds Dataset}

\author{ \href{https://orcid.org/0000-0002-0442-5474}{\includegraphics[scale=0.06]{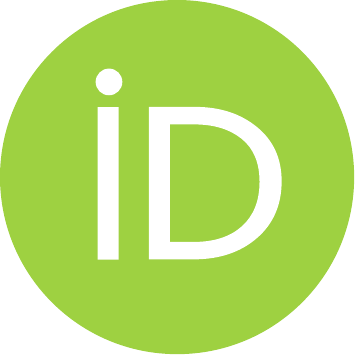}\hspace{1mm}Varun Aggarwal}\thanks{Use footnote for providing further
		information about author (webpage, alternative
		address)---\emph{not} for acknowledging funding agencies.} \\
	Electrical and Computer Engineering\\
	Purdue University\\
	West Lafayette, IN\\
	\texttt{aggarwal@purdue.edu} \\
	\And
    \href{https://orcid.org/0000-0002-7431-0808}{\includegraphics[scale=0.06]{orcid.pdf}\hspace{1mm}Aanis Ahmad} \\
	Electrical and Computer Engineering\\
	Purdue University\\
	West Lafayette, IN\\
	\texttt{ahmad31@purdue.edu} \\
	\And
	\href{https://orcid.org/0000-0003-2216-7875}{\includegraphics[scale=0.06]{orcid.pdf}\hspace{1mm}Aaron Etienne} \\
	Agricultural and Biological Engineering\\
	Purdue University\\
	West Lafayette, IN\\
	\texttt{aetienne@purdue.edu} \\
	\And
	\href{https://orcid.org/0000-0001-8882-0510}{\includegraphics[scale=0.06]{orcid.pdf}\hspace{1mm}Dharmendra Saraswat} \\
	Agricultural and Biological Engineering\\
	Purdue University\\
	West Lafayette, IN \\
	\texttt{saraswat@purdue.edu} \\

}

\renewcommand{\shorttitle}{\textit{arXiv} Template}

\hypersetup{
pdftitle={A template for the arxiv style},
pdfsubject={q-bio.NC, q-bio.QM},
pdfauthor={David S.~Hippocampus, Elias D.~Striatum},
pdfkeywords={First keyword, Second keyword, More},
}

\begin{document}
\maketitle

\begin{abstract}
Weeds are a major threat to crops and are responsible for reducing crop yield worldwide. To mitigate their negative effect, it is advantageous to accurately identify them early in the season to prevent their spread throughout the field. Traditionally, farmers rely on manually scouting fields and applying herbicides for different weeds. However, it is easy to confuse between crops with weeds during the early growth stages. Recently, deep learning-based weed identification has become popular as deep learning relies on convolutional neural networks that are capable of learning important distinguishable features between weeds and crops. However, training robust deep learning networks requires access to large imagery datasets. Therefore, an early-season weeds dataset was acquired under field conditions. The dataset consists of 159 Cocklebur images, 139 Foxtail images, 170 Redroot Pigweed images and 150 Giant Ragweed images corresponding to four common weed species found in corn and soybean production systems.. Bounding box annotations were created for each image to prepare the dataset for training both image classification and object detection deep learning networks capable of accurately locating and identifying weeds within corn and soybean fields. (\url{https://osf.io/w9v3j/})
\end{abstract}



\section{Introduction}
Weeds threaten crops worldwide causing considerable yield loss \citep{chauhan2020grand}. In the US corn belt alone, weeds have the potential to reduce corn and soybean crop yield by almost 50\% in fields without any weed control compared to the field with weed control \citep{Soltani2016, Soltani2017}. Therefore, it is crucial to control the spread of weeds within corn and soybean production systems. 

Implementing effective weed management practices is dependent on accurate weed identification. Currently, automation in weed identification has achieved high accuracy for identification due to the emergence of deep learning for computer vision. Although, training robust deep learning networks largely depends on the availability of high quality and well-annotated training data. Hence, this paper introduces a large annotated handheld imagery dataset called 4Weed Dataset. It consists of images acquired under complex field conditions for four different weed species: namely Cocklebur (\emph{xanthium strumarium}), Foxtail (\emph{setaria viridis}), Redroot Pigweed (\emph{amaranthus retroflexus}), and Giant Ragweed (\emph{setaria viridis}).

The 4Weed dataset was previously used in a study to accurately identify weeds using deep learning-based image classification and object detection networks. \citep{ahmad2021performance}.

\section{Materials and Methods}
\subsection{Dataset}

The 4Weed dataset consists of a total of 618 red, green, and blue (RGB) images that were acquired under complex field conditions at Purdue University's Agronomy Center for Research and Education (ACRE) during the summer months, as well as at the Purdue University greenhouse during winter months. The dataset is comprised of four common weed species to corn and soybean production systems: namely Cocklebur (\emph{xanthium strumarium}) images, Foxtail (\emph{setaria viridis}) images, Redroot Pigweed (\emph{amaranthus retroflexus}) images, and Giant Ragweed(\emph{setaria viridis}) (Figure \ref{figure:1}, Table \ref{tab:table}). The dataset can be found at: \url{https://osf.io/w9v3j/}

\begin{figure}[h]
\centering\includegraphics[width=0.9\linewidth]{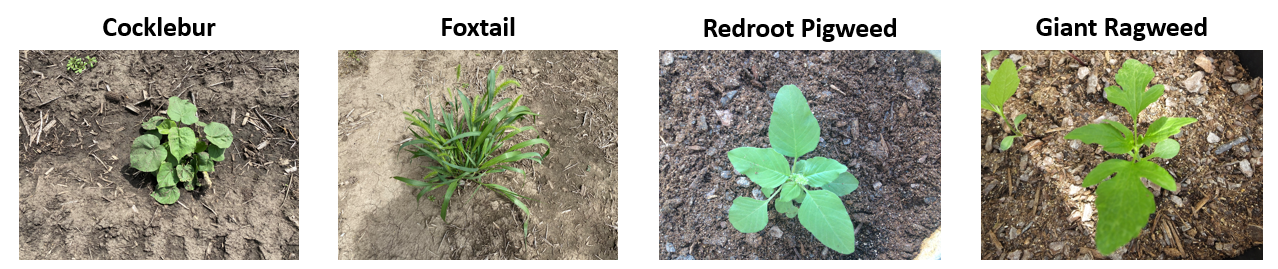}
\caption{Cocklebur, Foxtail, Redroot Pigweed, and Giant Ragweed images from 4Weed dataset}
\label{figure:1}
\end{figure}

\begin{table}[ht]
	\caption{4Weed Dataset Summary}
	\centering
	\begin{tabular}{lc}
		\toprule
		\cmidrule(r){1-2}
		Weed     & Number of Images \\
		\midrule
        Cocklebur          & 159\\ 
        Foxtail         & 139\\
        Redroot Pigweed & 170\\ 
        Giant Ragweed & 150\\ 
		\bottomrule
	\end{tabular}
	\label{tab:table}
\end{table}

The field images were collected using an array of six Logitech 920 web-cameras (Logitech, Newark, CA, USA) installed and positioned at the front end of a utility terrain vehicle (UTV) chassis. The resolution of the cameras was 1080 x 720 pixels. They were installed at a height of 14 inches above ground level (AGL). In addition, several images were captured at a resolution of 1200 x 900 pixels with a handheld Sony WX350 camera, both infield and greenhouse. Overall, the resulting dataset comprised of 35 Cocklebur images, 73 Foxtail images, 170 Redroot Pigweed images, and 150 Giant Ragweed images. 

As the initial dataset was unbalanced, additional images for Cocklebur and Foxtail were acquired using Google Pixel 3 and Apple iPhone 11 Pro mobile cameras in corn and soybean fields at Purdue University's ACRE farm. Each image had a resolution of 4032 x 3024 pixels. Hence, the final dataset was nearly balanced and had 618 images (Table \ref{tab:table}).

After the images were acquired, the LabelImg tool \citep{Tzutalin2015} annotation tool was used to create bounding box annotations and label them in order to prepare the dataset for training both image classification and object detection deep learning networks (Table \ref{tab:table}).

\clearpage

\bibliographystyle{model1-num-names}
\bibliography{references}

\end{document}